\documentclass[sigconf]{acmart}
\usepackage{multirow}
\usepackage{enumerate}
\usepackage{makecell}
\usepackage{bbm}
\usepackage{enumitem}
\usepackage{enumitem}
\usepackage{newtxtext,newtxmath}
\setlist{topsep=6pt, leftmargin=*}
\AtBeginDocument{%
  \providecommand\BibTeX{{%
    \normalfont B\kern-0.5em{\scshape i\kern-0.25em b}\kern-0.8em\TeX}}}

\setcopyright{acmcopyright}
\copyrightyear{2023}
\acmYear{2023}
\acmDOI{XXXXXXX.XXXXXXX}

\acmConference[Conference acronym 'XX]{Make sure to enter the correct
  conference title from your rights confirmation emai}{June 03--05,
  2018}{Woodstock, NY}
\acmPrice{15.00}
\acmISBN{978-1-4503-XXXX-X/18/06}

\definecolor{green}{rgb}{0, 0.5, 0}
\definecolor{orange}{rgb}{0.8, 0.6, 0.2}
\definecolor{orange2}{rgb}{1.0, 0.6, 0.2}
\definecolor{red}{rgb}{1.0, 0.0, 0.0}
\definecolor{teal}{rgb}{0.0, 0.4, 0.4}
\definecolor{purple}{rgb}{0.65,0,0.65}
\definecolor{saffron}{rgb}{0.95,0.75,0.2}
\definecolor{turquoise}{rgb}{0.0,0.5,0.5}
\definecolor{black}{rgb}{0.0, 0.0, 0.0}
\definecolor{gray}{rgb}{0.5, 0.5, 0.5}

\newcommand{\ying}[1]{{\color{black}#1}}
\newcommand{\li}[1]{{\color{red}#1}}

\begin{document}

\title{RetouchingFFHQ: A Large-scale Dataset for Fine-grained Face Retouching Detection}

\author{Qichao Ying}
\email{shinydotcom@163.com}
\affiliation{%
  \institution{Fudan University}
  \city{Shanghai}
  \country{China}
}

\author{Jiaxin Liu}
\email{19307130299@fudan.edu.cn}
\affiliation{%
  \institution{Fudan University}
  \city{Shanghai}
  \country{China}
}

\author{Sheng Li$^{\star}$}

\email{lisheng@fudan.edu.cn}
\affiliation{%
  \institution{NVIDIA}
  \city{Shanghai}
  \country{China}
}

\author{Haisheng Xu}
\email{hansonx@nvidia.com}
\affiliation{%
  \institution{NVIDIA}
  \city{Shanghai}
  \country{China}
}

\author{Zhenxing Qian}
\authornote{Corresponding authors: Sheng Li and Zhenxing Qian.}

\email{zxqian@fudan.edu.cn}
\affiliation{%
  \institution{Fudan University}
  \city{Shanghai}
  \country{China}
}

\author{Xinpeng Zhang}
\email{zhangxinpeng@fudan.edu.cn}
\affiliation{%
  \institution{Fudan University}
  \city{Shanghai}
  \country{China}
}

\fancyhead{}


\renewcommand{\shortauthors}{Ying et al.}

\begin{abstract}
The widespread use of face retouching filters on short-video platforms has raised concerns about the authenticity of digital appearances and the impact of deceptive advertising. To address these issues, there is a pressing need to develop advanced face retouching techniques. However, the lack of large-scale and fine-grained face retouching datasets has been a major obstacle to progress in this field. In this paper, we introduce RetouchingFFHQ
, a large-scale and fine-grained face retouching dataset that contains over half a million conditionally-retouched images. RetouchingFFHQ stands out from previous datasets due to its large scale, high quality, fine-grainedness, and customization. By including four typical types of face retouching operations and different retouching levels, we extend the binary face retouching detection into a fine-grained, multi-retouching type, and multi-retouching level estimation problem. Additionally, we propose a Multi-granularity Attention Module (MAM) as a plugin for CNN backbones for enhanced cross-scale representation learning. Extensive experiments using different baselines as well as our proposed method on RetouchingFFHQ show decent performance on face retouching detection. 

\end{abstract}


\begin{teaserfigure}
  \includegraphics[width=\textwidth]{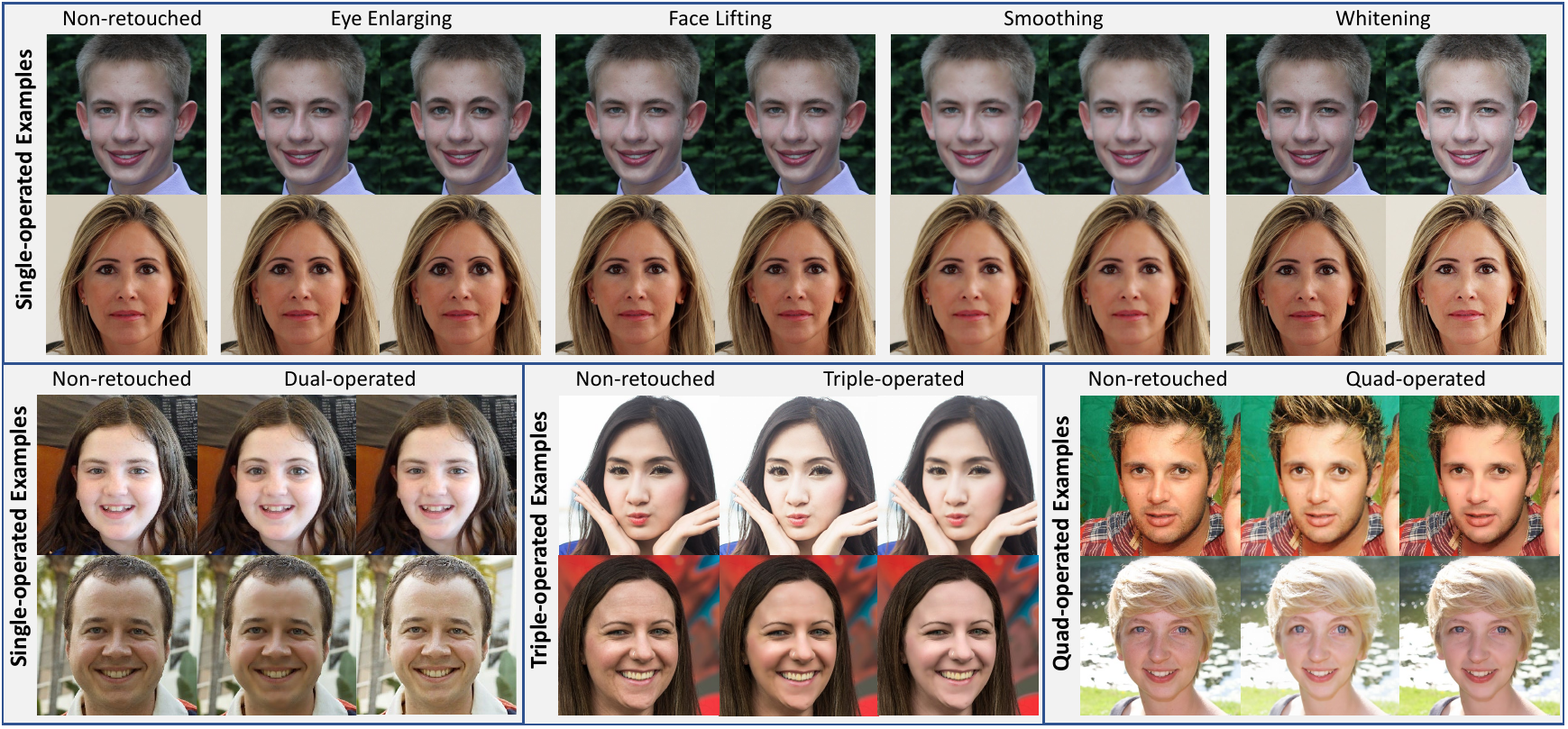}
  \caption{Examples of RetouchingFFHQ, a fine-grained face retouching dataset containing over half a million images. (Upper) Examples of single-operated images. In each operation group, the magnitude of the left / right images are respectively 30 / 90. (Lower) Examples of dual-operated to quad-operated images where the annotations are omitted for better visualization.}
  \label{fig:teaser}
\end{teaserfigure}


\maketitle

\section{Introduction}
The increasing popularity of short-video platforms such as TikTok, YouTube Shorts, and Facebook Reels has led to a surge in the use of face retouching filters~\cite{rathgeb2022handbook}. 
With just a click, users can achieve a more glamorous and visually striking appearance, thanks to a range of retouching functions such as face lifting, eye enlargement, whitening, and smoothing. 
While many people use these filters moderately and innocently, some go to the extreme to completely alter their appearance to deceive their audience for commercial purposes, e.g., news where people got scammed in online romance due to retouched appearance.

%

Currently, there are three main strategies to address the aforementioned issue. The first is \textbf{visible watermarking by law enforcement}.
For instance, the Norwegian government~\cite{Norwaylaw} 
has mandated that all ads featuring modified body shapes and facial appearances must be labeled with a visible disclaimer which takes up at least 7\% of the whole image. 
The United States, Israel and Germant also have similar labeling requirements~\cite{USlaw,Israellaw}.
The visible watermarks severely reduce the image quality, and it can deter users even from mild use of face retouching.
The second is \textbf{invisible watermarking}, which imperceptibly embeds fragile watermarks into the face images ~\cite{ying2021image,asnani2022proactive,he2020semi}. The watermarks will be damaged once the face images are retouched. This kind of approaches requires people to proactively embed the watermark into the face images once captured.
The third is \textbf{face image retouching detection} to determine whether the face images are retouched without any proactive processes. Face image retouching detection is a challenging problem as the retouching may only mildly change the image content. Directly applying the face manipulation (also DeepFake) detection~\cite{jain2018detecting,jain2020detecting} or image manipulation detection~\cite{MVSS,mantra} schemes would not be appropriate to tackle such a challenge. Be aware of this, researchers have devoted efforts to schemes which are developed specifically for face image retouching detection~\cite{wang2020deep,rathgeb2020differential}.  


\begin{table*}
\begin{center}
   \caption{A summary of existing datasets related to face retouching. RetouchingFFHQ is the largest and the first fine-grained dataset for face retouching consisting of 500k images processed by commercial APIs. $S$, $W$, $E$, $L$ respectively denote skin-smoothing, face-whitening, eye-enlarging and face-lifting.}
   \label{tab:dataset_compare}
\begin{tabular}{c|c|c|c|c|c}
\hline 
\multirow{2}{*}{Datasets} & \multicolumn{2}{c|}{Shots} & \multicolumn{2}{c|}{Image Attributes} & \multirow{2}{*}{Remarks}\\
\cline{2-5}
&  \#Ori. & \#Retou. & Operation & Level&\\
\hline
\hline
Kee et al.\cite{kee2011perceptual} & 468 & 468 & 1 (No distinction) & 5 (subjective) & \ying{Images crawed from the Internet}\\
\hline
YMU\cite{VMUYMU} & 151 & 300 & 1 (No distinction) & on / off  & Images of Caucasian females\\
\hline
VWU\cite{VMUYMU} & 51 & 153 & 3 (eyemakeup / lipstick / both) & on / off & Images of Caucasian females\\
\hline
Celebrity\cite{bharati2016detecting} & 165 & 165 & 1 (No distinction) & on / off & \ying{Images crawed from the Internet}\\
\hline
ND-IIITD\cite{bharati2016detecting} & 325 & 2275 & 7 (hybrid presets) & on / off & Using Portrait-pro Studio Max\\
\hline 
MDRF\cite{bharati2017demography} & 600 & 2400 & 2 (App type) & on / off & Using BeautyPlus/Potrait Pro\\
\hline
Rathgeb et al.\cite{rathgeb2020prnu} & 100 & 800 & 5 (hybrid filters) & on / off & Using five Apps with hybrid retouching\\
\hline
DiffRetouch\cite{rathgeb2020differential} & 1726 & 9078 & 7 (hybrid filters) & on / off & Improved version of Rathgeb et al.\cite{rathgeb2020prnu}\\
\hline
BeautyGAN~\cite{li2018beautygan} & 1115 & 2719 & 1 (No distinction) & on / off & GAN-based style-transfer with references\\
\hline
FFHQR\cite{shafaei2021autoretouch} & 70000 & 70000 & 1 (No distinction) & on / off & Hybridly retouched by professional editors \\
\hline
\hline
\textbf{RetouchingFFHQ} & 58158 & 652568 & \makecell{Single-operated ($S$, $W$, $E$, $L$)\\Dual-operated ($S+W$, etc.)\\Triple-operated ($S+W+E$, etc.)\\Quad-operated ($S+W+E+L$)} & \makecell{$S: 0/30/60/90$\\$W: 0/30/60/90$\\$E: 0/30/60/90$\\$L: 0/30/60/90$} & \makecell{Controllably Generated using Commercial\\APIs: Megvii, Alibaba and Tencent}\\
\hline 
\hline
\end{tabular}
\end{center}
\end{table*}

In literature, many datasets have been constructed for the task of face retouching detection, as summarized  in Table 1. 
Despite the progress, there are still several challenges that need to be addressed.
Firstly, the existing datasets are limited in size, with the largest one, i.e., FFHQR~\cite{shafaei2021autoretouch}, containing 70k retouched images, and the rest containing fewer than 10k samples.
Consequently, these datasets may not be sufficient to train robust detection networks.
Secondly, the current datasets lack fine-grained annotations of retouching types and levels.
For instance, some datasets, such as VWU~\cite{VMUYMU}, only contain images of Caucasian females with a limited range of retouching types. 
Other datasets, such as ND-IIITD~\cite{bharati2016detecting}, Rathgeb et al.\cite{rathgeb2019impact}, and DiffRetouch\cite{rathgeb2020differential}, employ binary classification for retouching detection, ignoring the diverse types and degrees of retouching. Kee et al.~\cite{kee2011perceptual} categorize the retouching levels, but still mix retouched images of various types. We believe that binary classification may not sufficient to capture the complexity of face retouching, which can range from innocuous cosmetic enhancement to malicious forgery. Therefore, there is a need to build more fine-grained face retouching datasets which can be used to train classifiers to capture the diverse types and levels of retouching.

To overcome the limitations of existing datasets, we construct a large-scale fine-grained face retouching dataset (RetouchingFFHQ) in this paper, which consists of over half a million retouched face images with a range of different retouching types and levels. Our dataset is constructed based on the well-known FFHQ dataset~\cite{FFHQ}, where different face retouching operations are performed on these images in terms of specific types and levels. To generate the retouched images, we utilize three popular commercial online APIs including Megvii~\cite{Megvii}, Tencent~\cite{Tencent}, and Alibaba~\cite{Alibaba}. Each retouched image in our dataset includes one to four types of retouching, namely, eye-enlarging, face lifting, skin smoothing, and face whitening, which are the most commonly used geometric and non-geometric operations on social media platforms. To ensure fine-grained annotations, we have categorized the levels of each retouching into four classes: off (0), slight (30), medium (60), and heavy (90). Figure 1 provides some examples, with more provided in the supplementary material.
The complete dataset will be made public for academic use after the anonymous reviewing stage.

Overall, the advantages of our dataset are listed below.

\begin{itemize}
\item \textbf{Large-Scality and High Quality}. 
RetouchingFFHQ is the largest face retouching dataset to date, consisting of 652,568 retouched images and 58,158 non-retouched images. To ensure high quality, we conducted strict data cleaning to exclude face images that might not be used in real-world face retouching applications, e.g., \textit{closed eyes}, \textit{poor lighting conditions}, \textit{incomplete faces}, etc.

\item \textbf{Fine-grained labeling.} RetouchingFFHQ extends the previous binary classification problem of face retouching detection into a fine-grained multi-label fashion. RetouchingFFHQ enables the detection of multiple types of retouching and their corresponding levels, providing a more nuanced understanding of the types and extent of face retouching in social media platforms.


\item \textbf{Cross APIs}. RetouchingFFHQ is composed of retouched images made by three commercial APIs. Users can evaluate the performance of the detection models using both in-API (training and testing based on the same API) and cross-API (training and testing based on different API) settings.
\end{itemize}

In addition to the dataset, we propose a novel Multi-granularity Attention Module (MAM) that can refine hierarchical representations and compare information from different receptive fields to improve the detection performance. Extensive experiments demonstrate that, with the proposed MAM as a plugin, the baselines can gain remarkable performance gain for face retouching detection.
With the proposed new dataset, we believe there is great potential for future work to tackle the challenging problem of real-world fine-grained face retouching detection.




\section{Related Works}
Table~\ref{tab:dataset_compare} summarizes the existing datasets related to face retouching.
Specifically, 
Kee et al.~\cite{kee2011perceptual} develop a dataset with 436 retouched images and invite volunteers to rank the amount of photo alteration on a scale of one to five.  
YMU and VMU~\cite{VMUYMU} are two compact datasets proposed to determine the presence of makeup operations from their impacts on automated face recognition.
Then, Bharati et al.\cite{bharati2016detecting} introduce two larger datasets, namely, ND-IIITD and Celebrity.
The ND-IIITD dataset contains seven presets, each containing eight to twelve different types of retouching. 
In~\cite{bharati2017demography}, Bharati et al further propose Multi-Demographic Retouched Faces (MDRF) to investigate the effect of demography and gender on the accuracy of retouching detection.
Rathgeb et al.\cite{rathgeb2020prnu} conduct qualitative assessment over beautification apps and five are chosen to create a new dataset. 
The photo response non-uniformity (PRNU) is used for retouching detection.
DiffRetouch~\cite{rathgeb2020differential} use seven apps, such as \textit{AirBrush} and \textit{BeautyPlus}, to generate nearly 10k hybrid retouched images, with each app showing different characteristics.
Apart from collecting data from apps, some recent methods also utilize GAN-based technologies to create new face retouching datasets.
BeautyGAN~\cite{li2018beautygan} build up a make-up dataset using style-transfer, which can achieve virtual makeup tryout.
Shafaei et al. propose FFHQR~\cite{shafaei2021autoretouch} which contains 70k retouched images by professional editors and is designed for automatic face retouching. 
Besides, some tailored detection networks~\cite{jain2018detecting,jain2020detecting,geng2020masked} are proposed for these datasets.


These datasets and detection networks are valuable and promote the development of the literature.
But there still exist issues regarding the binary-classification restriction and the limitation in dataset scale.
We attempt to address them via RetouchingFFHQ.

\begin{table*}
\begin{center}
   \caption{A Summary of RetouchingFFHQ, which contains three subsets with images retouched by different APIs.}
   \label{tab:dataset_summary}
\begin{tabular}{c|c|c|c|c|c}
\hline 
Image Kind & Available APIs & \#Retouched per sample & \#Num Retouched & \#Retouched/person & Avg. PSNR\\
\hline
\hline
Subset-0: Non-retouched& / &/&58158&/&/\\
\hline
\multirow{3}{*}{Subset-1: Single-operated}& Megvii &1 per \{sample, type\}&66940&4&29.33\\
&Tencent&1 per \{sample, type, level\}&200820&12&29.06\\
&Alibaba&1 per \{sample, type, level\}&19608&12&29.25\\
\hline
\multirow{2}{*}{Subset-2: Dual-operated}& Megvii &1 per \{sample, type\}&99972&6&28.75\\
&Tencent&1 per \{sample, type\}&99972&6&28.44\\
\hline
\multirow{2}{*}{Subset-3: Triple-operated}& Megvii &1 per \{sample, type\}&66228&4&28.26\\
&Tencent&1 per \{sample, type\}&66228&4&27.90\\
\hline
\multirow{2}{*}{Subset-4: Quad-operated}& Megvii &2 per \{sample, type\}&16400&2&27.82\\
&Tencent&2 per \{sample, type\}&16400&2&27.07\\
\hline 
\hline
\multicolumn{1}{c}{Total/Avg}&\multicolumn{1}{c}{}&\multicolumn{1}{c}{} &\multicolumn{1}{c}{710726}&\multicolumn{1}{c}{4.07}&\multicolumn{1}{c}{28.80} \\
\hline
\end{tabular}
\end{center}
\end{table*}
\begin{figure*}[!t]
	\centering
	\includegraphics[width=1.0\textwidth]{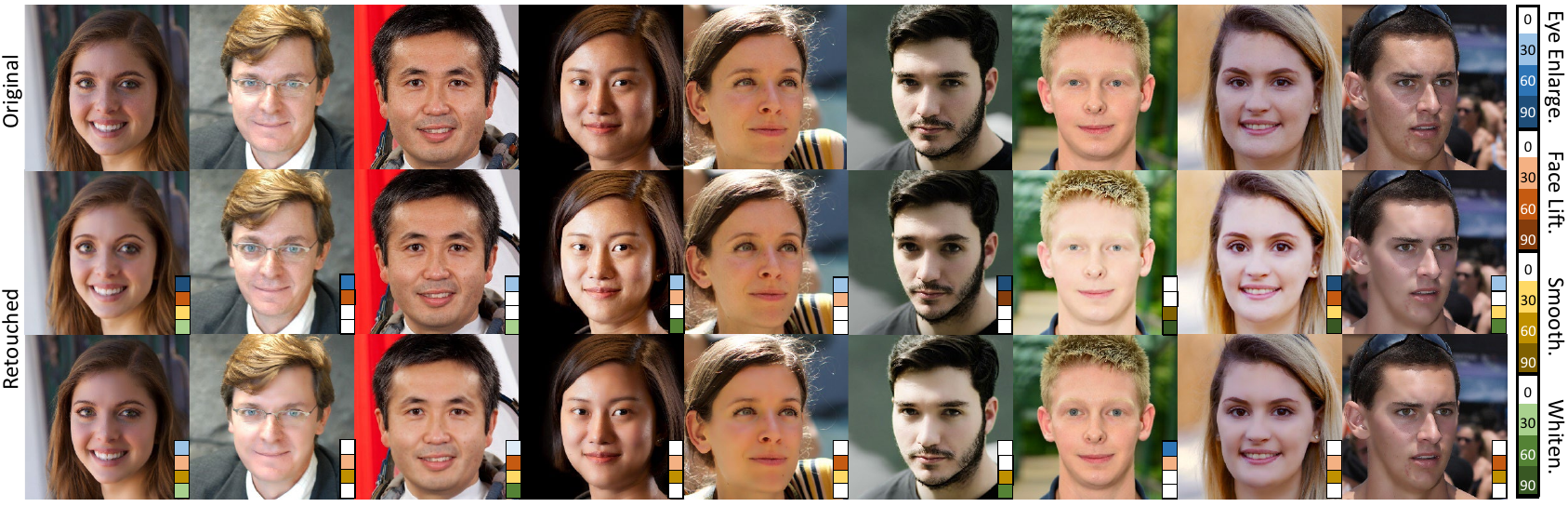}
	\caption{More examples of the dataset containing two to four operations. Labels are visualized in the downright corner.}
	\label{fig:more_example}
\end{figure*}

\section{RetouchingFFHQ}
This section introduces the collection and annotation process of RetouchingFFHQ, which is based on the Flicker-Faces-HQ dataset. 
The original FFHQ dataset contains 70,000 face-aligned images sized $1024\times~1024$ collected from the Flicker dataset~\cite{FFHQ}.
Due to its large quantity of high-quality face images and variety \& balance of ages, ethnicities, genders and backgrounds, we choose FFHQ as the source to build our new dataset.

\subsection{Data Collection and Cleaning}
\noindent\textbf{The Overall Strategy.}
To create our new dataset, we generate the fine-grained retouched images using the online commercial APIs of three famous AI companies including, Megvii~\cite{Megvii}, Alibaba~\cite{Alibaba} and Tencent~\cite{Tencent}.
For the selected APIs, we find that they all support four types of commonly-used retouching including, \textit{skin smoothing, face whitening, face lifting }and \textit{eye enlarging}.
Besides, the retouching levels of the three APIs range from zero to one or a hundred.
To quantify, we categorize the levels into four classes, namely, \textit{off} (0), \textit{slight} (30), \textit{medium} (60) and \textit{heavy} (90).
With the dataset and APIs, the collection of retouched images can be done via uploading images and queries containing type and level information.

\noindent\textbf{Face Image Cleaning.}
We find that a noticeable portion of images from FFHQ are either not proper to perform retouching, or occasionally appear in real-world face retouching scenarios. These images can be mostly summarized into the following five types.
1) Blurriness or Poor/Abnormal Lightning Condition, 
2) Cosplay, Already Filtered or with Heavy Makeups, 
3) Incomplete Faces,
4) Images of Infants and Babies, and
5) Fake Faces.
We manually check all the 70k images in FFHQ and remove the face images which belong to the five categorizes mentioned above. After the cleaning, we construct an original face image dataset with 58158 samples captured from 58158 different persons. 


\noindent\textbf{Retouched Face Image Collection.}
We split the cleaned original images into four non-overlapping subsets according to their indices in FFHQ, where images indexed 0-20k belong to the first subset, 20k-40k to the second, 40k-60k to the third, and 60k-70k to the last. 
We perform different amount of retouching operations for different original image subsets, the details of which are given below.
\begin{itemize}
\item \textbf{Subset-0} (\textit{Non-retouched images}): The subset is composed of the above cleaned images without any digital retouching.
\item \textbf{Subset-1} (\textit{Single-operated images}): Each original face image is individually retouched into four retouched versions by respectively applying one type of retouching, and the levels are evenly drawn from off (0), slight (30), medium (60) and heavy (90). Exampled images are \textit{skin-smoothed} images, \textit{eye-enlarged} images, etc. 
\item \textbf{Subset-2} (\textit{Dual-operated images}): Each original face image is individually retouched into six ($C_4^2$) retouched versions by respectively applying two type of retouching.
The levels of each performed retouching are also evenly and independently drawn from the three positive classes. Exampled images are \textit{skin-smoothed \& face-lifted} images, \textit{eye-enlarged \& face-whitened} images, etc. 
\item \textbf{Subset-3} (\textit{Triple-operated images}): Each original face image is individually retouched into four ($C_4^3$) retouched versions by respectively applying three type of retouching.
The levels are also evenly and independently drawn. Exampled images are \textit{skin-smoothed \& face-lifted \& eye-enlarged} images, etc. 
\item \textbf{Subset-4} (\textit{Quad-operated images}): Each original face image is retouched by applying all four retouching operations and the levels are evenly and independently drawn.
Because there are less original images in this subset, we generated two different retouched versions for each sample.
\end{itemize}

\begin{figure*}[!t]
	\centering
	\includegraphics[width=1.0\textwidth]{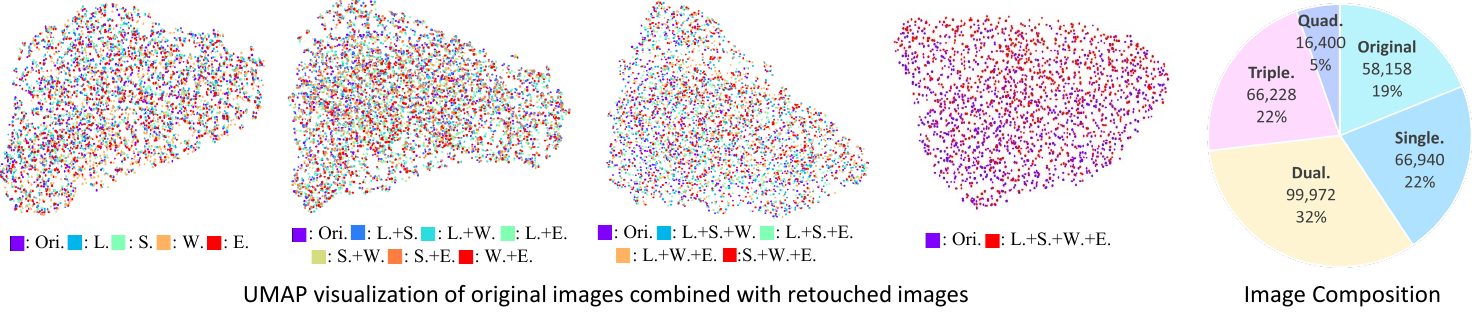}
	\caption{UMAP visualization and image composition of the Megvii retouching subset.}
	\label{fig:umap}
\end{figure*}
\begin{figure}[!t]
	\centering
	\includegraphics[width=0.48\textwidth]{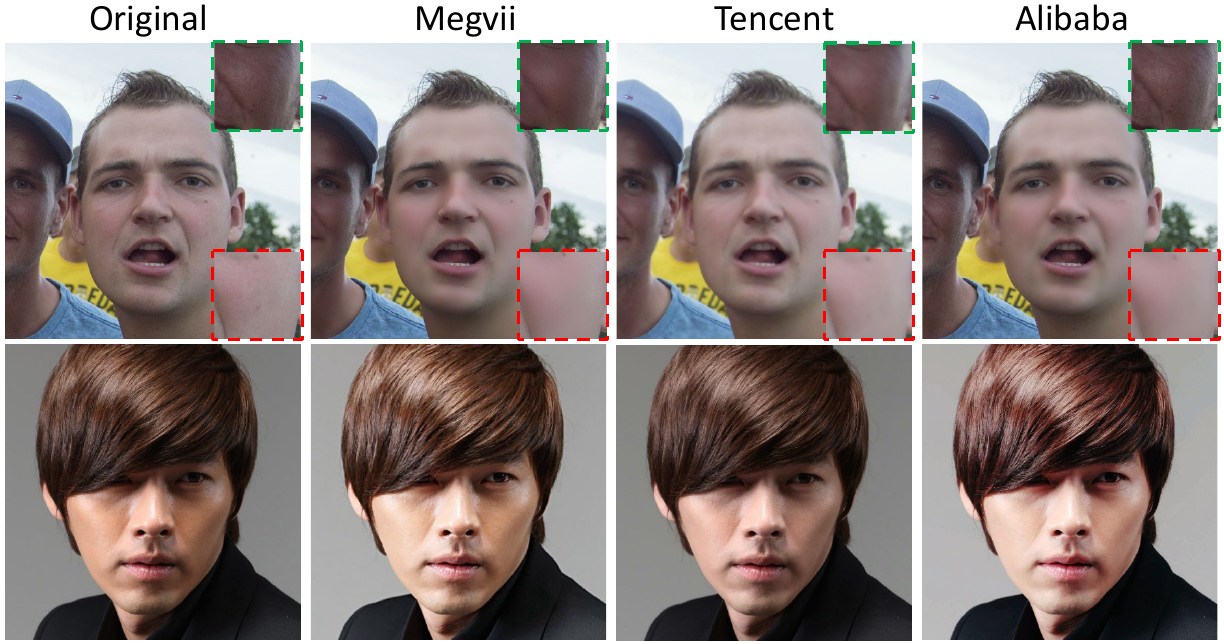}
	\caption{Examples of differences in visual effects among APIs. Annotations: (Upper) Smooth 90. (Lower) Smooth 90. Please zoom in to see differences in details and color.}
	\label{fig:API_diff}
\end{figure}
Besides the division according to the amount of retouching, RetouchingFFHQ can also be divided into three sets according to the source API of the images, namely, the \textbf{Megvii set}, the \textbf{Tencent set}, and the \textbf{Alibaba set}.
We save all the collected images in a uniform resolution $512\times~512$.
Notice that the Alibaba API only allow performing one retouching at a time, and thus we only collect the single-operated images in this subset.
Also, for Subset-1 of the Alibaba and Tencent set, we additionally provide images of each type in all three levels, resulting in two times more single-operated images.
Table~\ref{tab:dataset_summary} summarizes RetouchingFFHQ, which counts all subsets of the images reports the average PSNR between original images and their retouched versions w.r.t different APIs and the amounts of operations. Fig.~\ref{fig:more_example} showcase more examples of retouched faces randomly drawn from Subset 2-4, where the labels are attached in the downright.

\noindent\textbf{Data Splitting.}
We split the original face images (and their corresponding retouched versions) in our RetouchingFFHQ into training, validation and testing sets with a ratio of 80\%, 10\%, and 10\%. 


\subsection{Annotation and Objective Function}
\noindent\textbf{Annotation.} Each face image is annotated in the following form: 
\begin{equation}
    \{\texttt{Smooth}: s,\texttt{EyeEnlarge}: e,\texttt{FaceLift}: l,\texttt{Whiten}: 
    w\},
    \nonumber
\end{equation}
where $s$, $e$, $l$ or $w$ refer to the retouching level of each single operation.
We use zero to denote the ``off (0)" class, one to denote the ``slight (30)" class, two to denote the ``medium (60)" class, and three to denote the `` heavy (90)" class.
For instance, if a face image is retouched using medium eye-enlarging and heavy whitening. Its annotation would be therefore $
    \{\texttt{Smooth}: 0,\texttt{EyeEnlarge}: 2,\texttt{FaceLift}: 0,\texttt{Whiten}: 
    3\}$.

\noindent\textbf{Objective Function.}
For detection networks, we adopt an emsemble of four separate Multi-Layer Perceptrons (MLPs), where each MLP predicts the retouching level of a certain type. These MLPs are trained together in a multi-task manner. The overall loss function is an aggregation of four Cross-Entropy (CE) losses designed for the four MLPs, which is formulated in Eq.~(\ref{eqn:CE}).
\begin{equation}
\label{eqn:CE}
\mathcal{L}_{\emph{sum}}=\sum_{t\in\{S,E,L,W\}}\emph{CE}(\mathbf{Y}_{t},\mathbf{P}_{t})=-\sum_{t} \sum_{\emph{l}=0}^{3}y_{t,l}\log({p}_{t,l}),
\end{equation}
where $p_{t,l}$ is the predicted probability for type $t$ and level $l$, and $y_{t,l}$ is the ground-truth level. 

\subsection{Statistics and Visualizations}
\noindent\textbf{API Discrepancies.}
First, we find that the three APIs differ in the visual effects, as showcased in Fig.~\ref{fig:API_diff}.
It is mainly due to the characteristics of the underlying algorithms, as well as how levels correspond to the extent of modifications.
Therefore, besides mixing all retouched images from the three APIs as a whole to train joint models, we can also train them on the basis of a single API and conduct cross-API verification to test the generalization.

\noindent\textbf{Visualizations.}
We demonstrate the inter-category variability in Fig.~\ref{fig:umap} by randomly selecting 1000 original images and the corresponding retouching images from the Megvii dataset. 
We use a pretrained FaceNet model~\cite{facenet} to extract face representations and UMAP~\cite{UMAP} for visualization.
We see that the reduced features are mainly clustered with little inter-class discrepancies, suggesting that the retouching usually introduces moderate modification that poses challenges to face retouching detection. 




\section{The Multi-granularity Attention}
\label{section:MAM}
\noindent\textbf{Multi-granularity in Face Retouching Detection.}
We investigate how humans make predictions without reference to the original faces by scrutinizing the retouched images as shown in Fig.~\ref{fig:more_example}.
Besides geometric distortion or noise-level artifacts left by retouching algorithms,
we find that the other critical factor relies on the features that can be learnt from multiple granularities. For instance, given an image with large eyes shown in Fig.~\ref{fig:more_example} (row 1, column 1), a closer look on the eyes would easily lead to the conclusion that the image has undergone eye-enlarging. 
However, further considering the large occupation of the face in the image as well as the reasonable ratio of eyes and face, we would reconsider the image as not being eye-enlarged.
Similar phenomenon could also be observed on other retouching types.
Besides, there can exist a non-negligible amount of spatial redundancy within the visual representations. For example, the background and skin regions can be reduced into two tokenized representation containing the averaged statistic of lightning condition and sharpness.

\begin{figure}[!t]
	\centering	\includegraphics[width=0.48\textwidth]{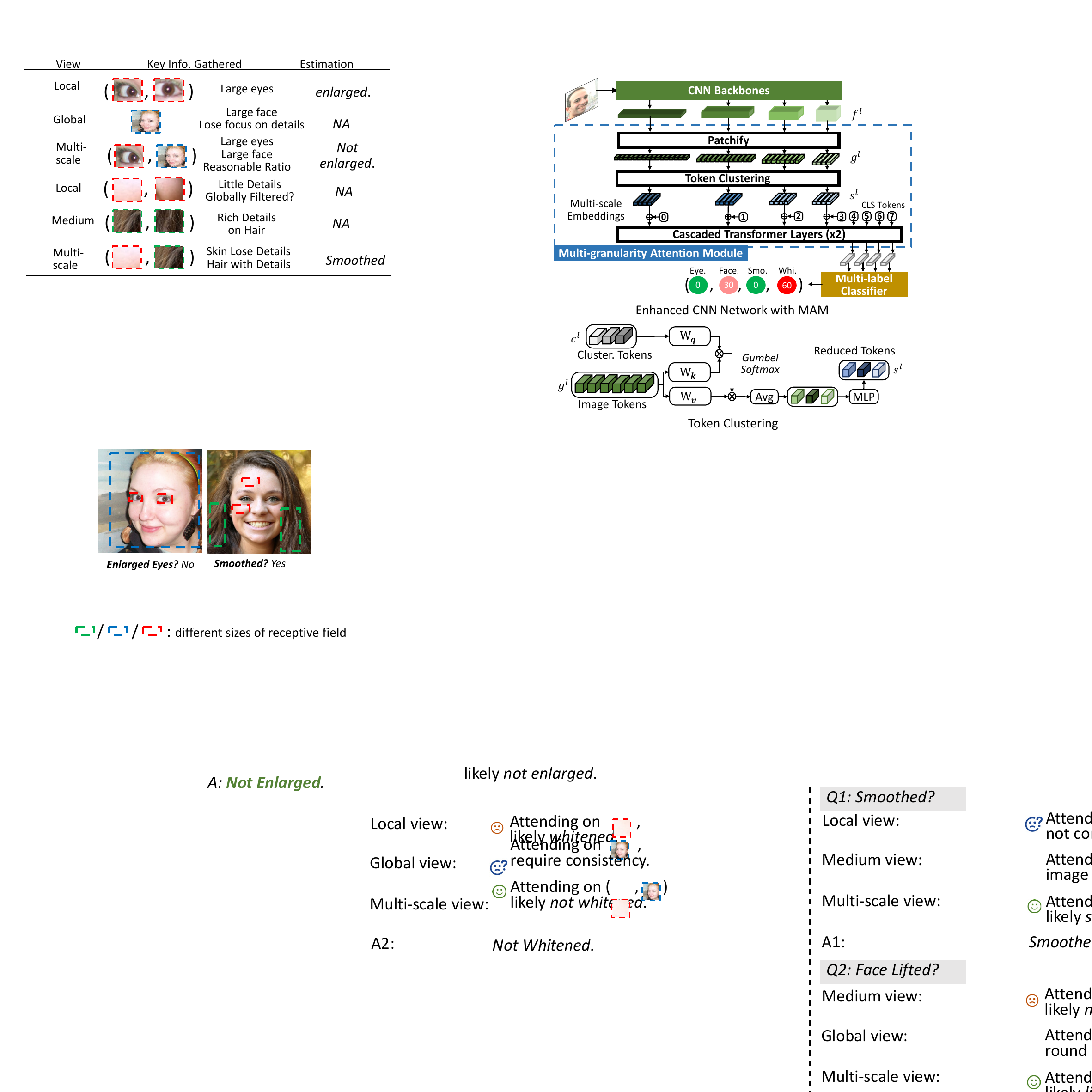}
	\caption{Network Design of the proposed MAM.}
	\label{fig:intuition}
\end{figure}
As shown in Fig.~\ref{fig:intuition}, for spatial redundancy reduction, we propose the adaptive token clustering method.
For enhanced multi-granularity representation learning, we employ a lightweight two-layered Transformer encoder~\cite{Vaswani2017attention} to analyze and compare multi-granularity information for detection. 


\noindent\textbf{Adaptive Token Clustering.}
We denote the hierarchical features from the CNN backbone as $F=\{f^0,...,f^{L-1}\}$ where $L$ denotes the amount of downsampling stages and usually equals four for typical CNN architectures.
Each $f^l, l\in[0,...,L-1]$ is generated by the last layer of each stage.
We patchify $f^l$ respectively into token representations~\cite{Swin} as $g^l=\{g_0^l,...,g_{HW/2^{2L}-1}^l\}$, where $H, W$ denote the height and width of the images, the window size is set as $1\times~1$ (each spatial measurement of $f^l$ corresponds to a separate token) and $i$ denotes the index of raster-scan ordering of each token.
Next, for each stage we introduce learnable \textit{clustering token embeddings} 
where the amounts $m=[m_0,...,m_{L-1}]=[\frac{HW/4}{r_0},...,\frac{HW/2^{2L}}{r_{L-1}}]$ are subject to empirically-set clustering rates $r=[r_0,...,r_{L-1}]$.
For each stage of the hierarchical CNN features, we assign each patchified tokens to one of the clustering token using the attention mechanism and the Gumble Softmax~\cite{Gumbel}.
\begin{equation}
\label{eqn:cluster}
    \mathbf{A}_{i, j}^l = \frac{\exp(W_Q{c}_i^{l} {\cdot} W_K{g}_j^l)}{\sum_{k=1}^{m_{l}} \exp(W_Q{c}_K^l{\cdot} W_K{g}_j^l)},
\end{equation}
where $j$ denotes the index of the clustering token, $W_Q$ and $W_K$ are the weights of the learned linear projections for \textit{keys} and \textit{queries} of the hierarchical image tokens and the clustering tokens, respectively.
$\mathbf{A}^l$ is the attention matrix that records the pseudo-one-hot value of assignment of each image token to a single cluster, with gradient enabled.
Afterward, we combine the embedding of all the image tokens belonging to the same group by taking the average of the projected representations for spatial redundancy reduction. 
Correspondingly, each clustering token ${{c}}_i^{l}$ corresponds to a new reduced tokens ${{s}}_i^{l}$, and we further linearly project all ${{s}}_i^{l}$ using a shared single-layered MLP.
\begin{equation}
\label{eqn:mapping}
    {s}_i^{l} =  W_o\frac{\sum_{j=1}^{M_{l-1}}\mathbf{A}_{i, j}^{l}W_V{g}_j^l}{\sum_{j=1}^{M_{l-1}} \mathbf{A}_{i, j}^{l}},
\end{equation}
where $W_V$ and $W_{o}$ are the MLPS for representation projections.
The reduced tokens for each stage is represented as $s^l=\{s_0^l,...,s_{m_l}^l\}$


\noindent\textbf{Multi-granularity Attention.}
Afterwards, we concatenate the hierarchy of the reduced tokens $s^l$, append four learnable tokens as the \texttt{CLS} tokens, add multi-scale positional embeddings where token from the same level share a same embedding, and feed them into two (denoted as $d=2$) cascaded layers of transformer encoders.
The correspondings representations at the four \texttt{CLS} tokens are then separately fed into the multi-labeled classifiers, which separately predict the levels of the four retouching operations.

\begin{table*}
\begin{center}
   \caption{In-API performance on the Megvii set. Ratio of single-operated to quad-operated images are shown in Fig.~\ref{fig:umap}.}
   \label{table:full_experiment_megvii}
\begin{tabular}{c|ccc|ccc|ccc|ccc|ccc}
\hline 
\multirow{2}{*}{Network}&\multicolumn{3}{c}{Eye Enlarging} &\multicolumn{3}{c}{Face Lifting}&\multicolumn{3}{c}{Smoothing} &\multicolumn{3}{c}{Whitening} &\multicolumn{3}{c}{Sum}\\\
& TP & TN & AC & TP & TN & AC & TP & TN & AC & TP & TN & AC & TP & TN & AC
\\
\hline  \hline
VGG16~\cite{VGG}&.0 & 1. & .567 & .0 & 1. & .567 & .977 & .991 & .933 & .761 & .965 & .782 & .164 & .974 & .285\\
InceptionV3~\cite{inceptionv2} &.689 &.917 &.658 &.451 &.824 &.554 &.851 &.996 &.770 &.862 &.780 &.700 &.502 &.765 &.266\\
ResNet50~\cite{ResNetv2} &.562 &.878 &.610 &.037 &.984 &.564 &.880 &.996 &.871 &.804 &.842 &.712 &.304 &.833 &.260\\
ConvNextV1~\cite{ConvNext} &.627 &.904 &.626 &.331 &.894 &.571 &.895 &.984 &.853 &.811 &.848 &.748 &.440 &.815 &.285\\
DenseNet121~\cite{DenseNet} &.681 &.924 &.668 &.525 &.805 &.559 &.915 &.986 &.831&.864&.805&.736&.546&.755&.281\\
EfficientNet~\cite{tan2019efficientnet} &.504 &.967 &.646 &.122 &.974 &.576 &.921 &.937 &.800 &.689 &.963 &.738 &.316 &.909 &.283\\
\hline 
\hline
Swin-T~\cite{Swin} &.0&1.&.567&.0&1.&.567&.0&1.&.567&.637&.925&.572&.043&.955&.180\\
CoAtNet~\cite{dai2021coatnet} &.0&1.&.567&.0&1.&.567&.879&.983&.684&.692&.951&.602&.144&.961&.206\\
CrossViT~\cite{chen2021crossvit}&.0&1.&.567&.0&1.&.567&.221&.851&.511&.366&.978&.573&.046&.900&.163\\
Conformer~\cite{conformer} &.616 &.849 &.559 &.004 &.998 &.568 &.961 &.932 &.656 &.768 &.954 &.608 &.309 &.850 &.201\\
\hline 
\hline
\textbf{InceptionV3-MAM} &.634 &.938 &.666 &.381 &.882 &.570 &.983 &.990 &.901 &.923 &.877 &.808 &.514\li{+.012} &.825 &.325\li{+.059}\\
\textbf{ResNet50-MAM} &.535 &.945 &.647 &.212 &.929 &.566 &.959 &.995 &.919 &.861 &.953 &.832 &.395\li{+.091} &.897 &.330\li{+.070}\\
\textbf{DenseNet121-MAM} &.735 &.913 &.679 &.568 &.856 &.598 &.987 &.979 &.916 &.829 &.965 &.825 &.598\li{+.052} &.835 &.356\li{+.075}\\
\hline 
\hline
\end{tabular}
\end{center}
\end{table*}
\begin{table}
\begin{center}
   \caption{In-API performance on the triple- and quad- operated images of the Megvii set.}
   \label{table:multi_process_megvii}
\begin{tabular}{c|c|c|c|c|c}
\hline 
\multirow{2}{*}{Network}&{Eye} &{Face}&{Smo.} &{Whi.} &{Sum}\\\
& TP & TP & TP &  TP & TP
\\
\hline  \hline
InceptionV3~\cite{inceptionv2}&.634&.184&.841&.801&.154 \\
ResNet50~\cite{ResNetv2}&.490&.310&.496&.545&.078\\
ConvNextV1~\cite{ConvNext}&\textbf{.836}&.156&.715&.651&.111\\
DenseNet121~\cite{DenseNet}&.624&.260&.889&.844&.205 \\
EfficientNet~\cite{tan2019efficientnet}&.503&.008&.922&.683&.077\\
\hline 
\hline
\textbf{InceptionV3-MAM}&.539&.133&.976&\textbf{.918}&.171~\li{+.017}\\
\textbf{ResNet50-MAM}&.510&.008&.917&.863&.086~\li{+.009}\\
\textbf{DenseNet121-MAM}&.695&\textbf{.442}&\textbf{.986}&.821&\textbf{.350}~\li{+.145} \\
\hline 
\hline
\end{tabular}
\end{center}
\end{table}

\section{Experiments and Analysis}
\label{section:experiment}
\subsection{Settings and Performance Indicators}
\noindent\textbf{Settings.}
We employ a number of representative CNN based, Transformer based and hybrid deep network architectures as the baselines,
namely, VGG-Net~\cite{VGG}, ResNet50~\cite{ResNetv2}, DenseNet121~\cite{DenseNet}, InceptionNetV3~\cite{inceptionv2}, ConvNextV1~\cite{ConvNext}, EfficientNet~\cite{tan2019efficientnet},CrossViT~\cite{chen2021crossvit}, Swin-T~\cite{Swin}, Conformer~\cite{conformer} and CoAtNet~\cite{dai2021coatnet}.
To show the effectiveness of MAM, we apply them into the InceptionNet, DenseNet and ResNet as examples.
We train all networks with eight as the default batch size, Adam as the default optimizer~\cite{kingma2014adam} and $2\times10^{-4}$ as the learning rate for all CNN-based layers and $1\times10^{-4}$ for all Transformer-based layers. 
The networks are trained till converge for 20-30 epochs with adaptive early exiting.
Each model is separately trained on a NVIDIA RTX 3090 GPU.
Though some related works~\cite{FFHQWrinkle,shafaei2021autoretouch} have proposed their benchmarks for retouching detection, they are verified on either smaller or binary-classification datasets.
Besides, recent architectures such as ConvNext or Conformer contain more modern designs.
For MAM, the clustering rate is set as $r=[\frac{1}{64},\frac{1}{16},\frac{1}{4},1]$ by default for each level. If $r_i=1$ for any stage, we skip the clustering function in Eq.~(\ref{eqn:cluster}) and only conduct the mapping function in Eq.~(\ref{eqn:mapping}).

In the below experiments, we use all four subsets of each API set to train unified models for detecting images with different amounts of operations.
Also, during training, we optionally perform reduced sampling (33.3\%) on the original images in all sets and on the single-operated images in Tencent and Alibaba dataset, to prevent over-performing on non- or slightly-retouched images.
The Alibaba set is reserved only for cross-API performance validation.

\noindent\textbf{Performance Indicators.}
\label{section:indicator}
We define the true positive, true negative and absolute correctness on evaluating algorithms as follows:
\begin{itemize}
\item \textbf{True Positive}, or \emph{TP}, which regulates that the detection algorithm should correctly figure out which types of retouching are \textit{on}. Given an image, \emph{TP} holds only if the algorithm predicts positive (non-zero class) on \textit{all} the performed operations.
The indicator does not take into account the correctness of predictions on specific levels or the not-performed operations.
\item \textbf{True Negative}, or \emph{TN}, which regulates that the detection algorithm should correctly figure out which types of retouching are \textit{off}. Given an image, \emph{TN} holds only if the algorithm predicts negative (zero class) on \textit{all} the not-performed operations.
It does not take into account the correctness of predictions on the performed operations.
\item \textbf{Absolute Correctness}, or \textit{AC}, which regulates that the algorithm must accurately predict the level of each face retouching, no matter being \textit{on} or \textit{off}. Given an image, \emph{AC} holds only if the prediction exactly matches the annotation.
\end{itemize}

Besides, the three indicators can be also calculated based on the predictions of each retouching operation, by ignoring the accuracies of the rest operations. 
The mathematical definitions of the indicators are as follows. 
\begin{equation}
\label{eqn_eval}
\begin{gathered}
\emph{TN}_{\emph{type}}=\frac{\mathbbm{1}(y_{\emph{type}}=\hat{y}_\emph{type}=0)}{\mathbbm{1}(y_{\emph{type}}=0)},
\emph{TN}_{\emph{sum}}=\frac{\mathbbm{1}(\cap_{\emph{type}}y_{\emph{type}}=\hat{y}_\emph{type}=0)}{\mathbbm{1}(\cup_{\emph{type}}y_{\emph{type}}=0)},\\
\emph{TP}_{\emph{type}}=\frac{\mathbbm{1}(y_{\emph{type}}=\hat{y}_\emph{type}\neq~0)}{\mathbbm{1}(y_{\emph{type}}\neq~0)},
\emph{TP}_{\emph{sum}}=\frac{\mathbbm{1}(\cap_{\emph{type}}y_{\emph{type}}=\hat{y}_\emph{type}\neq~0)}{\mathbbm{1}(\cup_{\emph{type}}y_{\emph{type}\neq~0)}},\\
\emph{AC}_{\emph{type}}=\frac{\mathbbm{1}(y_{\emph{type}}=\hat{y}_\emph{type})}{\mathbbm{1}(y_{\emph{type}})},
\emph{AC}_{\emph{sum}}=\frac{\mathbbm{1}(\cap_{\emph{type}}y_{\emph{type}}=\hat{y}_\emph{type})}{\mathbbm{1}(\cup_{\emph{type}}y_{\emph{type}})},
\end{gathered}
\end{equation}
where $y$ and $\hat{y}$ respectively denote the ground-truth level and the predicted level that takes the argmax of the network output. 
$\mathbbm{1}(x)$ counts the amount of items that let $x=\texttt{True}$.  

\noindent\textbf{Data Augmentation for Performance Analysis.}
Owing to the fact that retouched images are often lossily compressed during data transmission, we encourage users perform random lossy operations on the retouched images both in the training and testing stage.
Specifically, we randomly perform motion blurring using the PyTorch Albumentations library~\cite{Albumentations} with kernel size ranging from three to seven and 50\% activation possibility (\texttt{p=0.5}), and save the images 50\% possibility in PNG format and another 50\% possibility in JPEG format whose quality factors evenly range from 80 to 95.
We report the averaged performance by testing five times to measure the averaged performances against lossy operation attacks.
\begin{equation}
\label{eqn:average}
\emph{TP}=\sum_{i=1}^5{\emph{TP}^i}, \emph{TN}=\sum_{i=1}^5{\emph{TN}^i}, \emph{AC}=\sum_{i=1}^5{\emph{AC}^i},
\end{equation}
where the superscripts denotes different tests.
Though this would introduce randomness, due to the large scale of the dataset, we find that the resultant performance discrepancies among different tests are mostly not significant, with $\triangle\emph{TP}\leq~1\%$. The following experiments are reported in this form.





\begin{table*}
\begin{center}
   \caption{Cross-API performance. Models trained on the Megvii set are tested on the Tencent set. Since different APIs determine levels according to their own criteria, we only evaluate TP in the test.}
   \label{table:cross_tencent}
\begin{tabular}{c|c|ccc|ccc|ccc|ccc|ccc}
\hline 
&\multirow{2}{*}{Network}&\multicolumn{3}{c}{Eye Enlarging} &\multicolumn{3}{c}{Face Lifting}&\multicolumn{3}{c}{Smoothing} &\multicolumn{3}{c}{Whitening} &\multicolumn{3}{c}{Sum}\\\
& & TP & TN & AC & TP & TN & AC & TP & TN & AC & TP & TN & AC & TP & TN & AC
\\
\hline  \hline
\multirow{3}{*}{\rotatebox{90}{Direct}} & DenseNet121-MAM &.783 &.903 &.698 &.376 &.962 &.668 &.086 &.989 &.638 &.041 &.974 &.621 &.196 &.892 &.185\\
& InceptionV3-MAM &.619 &.961 &.673 &.112 &.976 &.629 &.017 &.988 &.632 &.115 &.906 &.587 &.119 &.895 &.150\\
& ResNet50-MAM &.559 &.956 &.665 &.114 &.974 &.628 &.005 &.997 &.634 &.048 &.963 &.616 &.100 &.930 &.150\\
\hline  \hline
\multirow{3}{*}{\rotatebox{90}{Finetune}} & DenseNet121-MAM &.918 &.995 &.938 &.930 &.987 &.915 &.999 &.951 &.968 &.911 &.994 &.856 &.907 &.953 &.722\\
& InceptionV3-MAM &.907 &.942 &.840 &.642 &.989 &.761 &.997 &.935 &.958 &.950 &.973 &.835 &.803 &.900 &.530\\
& ResNet50-MAM &.670 &.998 &.840 &.413 &.985 &.709 &.999 &.884 &.926 &.784 &.998 &.830 &.601 &.914 &.477\\
\hline  \hline
\multirow{3}{*}{\rotatebox{90}{Full}} & DenseNet121-MAM &.940 &.994 &.947 &.899 &.998 &.927 &.997 &.987 &.991 &.984 &.983 &.893 &.931 &.976 &.788\\
& InceptionV3-MAM &.868 &.995 &.913 &.755 &.980 &.791 &.999 &.952 &.970 &.942 &.996 &.905 &.835 &.951 &.651\\
& ResNet50-MAM &.892 &.974 &.879 &.825 &.970 &.819 &.999 &.973 &.983 &.981 &.981 &.894 &.883 &.936 &.650\\
\hline 
\hline
\end{tabular}
\end{center}
\end{table*}

\begin{table}[!t]
\centering
  \caption{Ablation studies and parameter settings. Network: DenseNet121-MAM. Task: in-API testing on Megvii.}
\label{table_ablation}
\begin{tabular}{c|ccc}
\hline
\multirow{2}{*}{Test} &\multicolumn{3}{c}{Sum} \\
  \cline{2-4}  & TP & TN & AC \\

\hline
\hline
Smaller clustering rate: $r=[\frac{1}{32},\frac{1}{8},\frac{1}{2},1]$ & .397 & .708 & .192 \\
Larger clustering rate: $r=[\frac{1}{128},\frac{1}{32},\frac{1}{8},\frac{1}{2}]$ & .355 & .842 & .230 \\
More trans. layers within MAM: $d=4$ & .473 & .902 & .347\\
Less trans. layers within MAM: $d=1$ &.425&\textbf{.919}&.331\\
without data cleaning & .342 & .877 & .257\\
without data augmentation & .279 & .854 & .175\\
\textbf{Default Setting}  & \textbf{.519}  & .895 & \textbf{.366} \\
\hline
\end{tabular}
\end{table}

\begin{table}
\begin{center}
   \caption{Cross-API performance. Models trained on the Megvii set are tested on the Alibaba set (containing only single-operated images) directly, without any finetuning.}
   \label{table:cross_alibaba}
\begin{tabular}{c|c|c|c|c|c}
\hline 
\multirow{2}{*}{Network}&{Eye} &{Face}&{Smo.} &{White.} &{Sum}\\\
& TP & TP & TP &  TP & TP
\\
\hline  \hline
InceptionV3~\cite{inceptionv2}&.430&.467&.633&.605&.534 \\
ResNet50~\cite{ResNetv2}&.513&.449&.373&.496&.458\\
ConvNextV1~\cite{ConvNext}&.721&.436&.557&.602&.579 \\
DenseNet121~\cite{DenseNet}&.526&.553&.662&.636&.594 \\
EfficientNet~\cite{tan2019efficientnet}&.326&.169&.771&.380&.412 \\
\hline 
\hline
\textbf{InceptionV3-MAM}&.440&.368&.795&.555&.539~\li{+.005}\\
\textbf{ResNet50-MAM}&.390&.302&.668&.505&.466~\li{+.008}\\
\textbf{DenseNet121-MAM}&.519&.548&.861&.501&.607~\li{+.013} \\
\hline 
\hline
\end{tabular}
\end{center}
\end{table}

\subsection{In-API Performances}
We show the performance on the Megvii set in Table~\ref{table:full_experiment_megvii} and that on the Tencent set in Table~\ref{table:cross_tencent}. We also conduct ablation studies on the in-API Megvii set test in Table~\ref{table_ablation}.
We find that CNN architectures can perform generally better in our task compared to the transformer and hybrid counterparts.
For example, the overall TP for InceptionNet and DenseNet are above 0.5, which leads by a noticeable margin.
Yet we see that for these models as well as many of the rest, the detection accuracies on geometric operations are lacking.
We introduce MAM into respectively InceptionNet, ResNet and DenseNet, and find that the overall performance w.r.t both TP and AC increased a lot, and the detection results on the geometric operations are significantly improved.
Our view is that most CNN backbones have their classifiers operated on the lowest-resolutioned representations, e.g., $7\times~7$, which might lack multi-granularity modeling.
Though the residual connection~\cite{ResNetv2} and inception-like modules~\cite{inceptionv2} can combine features of different receptive fields, the operations are on the basis of whole feature maps, therefore retaining spatial redundancy and lacking focus on what to attend. 
In Table~\ref{table:multi_process_megvii}, we further report the TPs on the comparatively heavy retouched, i.e., triple- and quad- operated images.
For all four operations, after adding the MAM on different backbones, the TPs are mostly improved.

\noindent\textbf{Ablation Studies.}
We also study the effect of data cleaning and network designs of MAM.
Notice that in ``w/o data cleaning", we exclude invalid data in the test set, meaning that there will be more but noisier data in the training stage.
In ``w/o data augmentation", we train the models on the unaugmented images and test them on the augmented images.
The studies show that the existence of noisy data and excessive image details will mislead the models from learning faulty representations, and the proper setting of $r$ and $l$ also affects the overall performance.

\subsection{Cross-API Performances}
Due to that AC is strict in definition which does not consider API discrepancies, we report the TP values in each test.
Provided with the models trained on the Megvii set, we empirically find that they can be directly tested on the Alibaba set, while it requires some finetuning to be well-roundedly tested on the Tencent set.
We finetune the models on the Tencent set for 10,000 iterations (accounting for around 15\% of all images) and report the performances together in Table~\ref{table:cross_tencent}. 
For reference, we also train models from scratch till convergence on the Tencent set, and report the in-API performance in the same table.
We find that the short finetuning is effective enough for models to adjust to the new protocols of the new platform.
The performance of training on the Tencent set and testing on the Megvii set yields similar results.
We omit it for space limit.
Table~\ref{table:cross_alibaba} reports the performances by direct cross-API application on the Alibaba set.
The results show that both the baseline networks and those with MAM have performance generalization. 
While the performances of models with MAM on in-API tests are better, it is also the case for cross-API tests w.r.t. overall TPs.

\section{Conclusion}
\label{section:conclusion}
In this paper, we introduce a large-scale and fine-grained face retouching dataset, RetouchingFFHQ, which enables more effective research in the field of face retouching detection. By considering different retouching types and levels, we extend the binary detection problem into a fine-grained, multi-retouching type, and multi-retouching level estimation problem. To enhance feature extraction, we propose a Multi-granularity Attention Module (MAM) as a plugin for CNN backbones. Extensive experiments using different baselines as well as our proposed method on RetouchingFFHQ show decent performance on face retouching detection.


\bibliographystyle{ACM-Reference-Format}
\balance
\bibliography{sample-sigconf}

\end{document}